\documentclass[journal]{IEEEtran}
\usepackage{mathtools}
	
\usepackage{xcolor}

\usepackage{color,soul}
\usepackage{pbox}

\usepackage{comment}
\usepackage{soul,framed,caption}
\usepackage[english]{babel}
\usepackage[utf8]{inputenc}

\usepackage[noend]{algpseudocode}
\usepackage{ifpdf}
\usepackage{float}
\usepackage{graphicx}
\usepackage{multirow}

\usepackage{enumerate}
\usepackage{algorithm}
\usepackage{graphicx}
\usepackage{float}
\usepackage{cite}
\usepackage{graphicx}
\usepackage{multirow}
\usepackage{array}
\usepackage{amssymb}
\usepackage{url}
\usepackage{makecell}
\usepackage{cite}
\usepackage{textcomp}

\usepackage{csquotes}
\usepackage{tikz}
\usepackage{tikz}
\usetikzlibrary{calc}
\makeatletter
\newlength{\mylength}

\newsavebox{\mybox}

\usepackage{nomencl}
\makenomenclature
 
\ifCLASSINFOpdf
  
\else
 
\fi

\usepackage{amsmath}

\usepackage{array}
\usepackage{subcaption}
 \usepackage{dblfloatfix}
\usepackage[figurename=Fig.,font=footnotesize]{caption}
\usepackage{url}

\begin{document}
\title{On Designing Features for Condition Monitoring of Rotating Machines}
 \author{Seetaram Maurya and Nishchal K. Verma, \textit{Senior Member, IEEE}
 \thanks{S. Maurya and N. K. Verma are with the Indian Institute
of Technology Kanpur, U.P. 208016, India (e-mail: seetaramiitk@gmail.com; nishchal@iitk.ac.in).}
 }
\maketitle

\begin{abstract}
Various methods for designing input features have been proposed for fault recognition in rotating machines using one-dimensional raw sensor data. The available methods are complex, rely on empirical approaches, and may differ depending on the condition monitoring data used. Therefore, this article proposes a novel algorithm to design input features that unifies the feature extraction process for different time-series sensor data. This new insight for designing/extracting input features is obtained through the lens of histogram theory. The proposed algorithm extracts discriminative input features, which are suitable for a simple classifier to deep neural network-based classifiers. The designed input features are given as input to the classifier with end-to-end training in a single framework for machine conditions recognition. The proposed scheme has been validated through three real-time datasets: a) acoustic dataset, b) CWRU vibration dataset, and c) IMS vibration dataset. The real-time results and comparative study show the effectiveness of the proposed scheme for the prediction of the machine's health states.
\end{abstract}

\IEEEpeerreviewmaketitle

\section{Introduction}
\label{Introduction}
Condition-Based Monitoring (CBM) schemes are highly practiced for rotating machinery components. The machines operate in highly demanding conditions with high structural vibration, rotation, temperature, pressure, etc. Therefore, over time, the components of rotary machines deviate from their healthy state to different faulty states. It is vital to apply the CBM system to identify the current faulty states for optimal maintenance action. In industries, the CBM systems also help in improving security, components' lifetime, and total productivity \mbox{\cite{intro22}}. CBM systems can be developed using different sensor signals such as vibration, acoustic, temperature,  etc. \cite{intro01, intro11}. The condition monitoring can be performed using model-based methods, signal-based methods, and Machine Learning (ML)-based methods \cite{ intro09, intro10, seetaramtim, intro12}. 
The model-based CBM schemes require mathematical models of rotating systems available either by using physics or system identification methods. Whereas signal processing and ML-based CBM schemes usually involve four important stages: a) data acquisition: collection of data about the different states of the machine, b) pre-processing: removal of outliers and redundant data, c) data segmentation and/or feature extraction: obtaining set of input features, and d) classification: recognition of health states of the machine. The performance of the classifiers is highly dependent on the quality and quantity of input features \mbox{\cite{intro09,intronew1}}. Therefore, an appropriate set of input features must be designed for the data-driven CBM schemes.

In recent literature, ML methods, such as multi-layer Neural Networks (NN), Random Forest (RF), Support Vector Machine (SVM), Convolutional Neural Network (CNN), Recurrent Neural Network (RNN), Deep Neural Network (DNN) models, etc., have been widely practiced to develop CBM schemes \mbox{\cite{intro09}}. These ML methods need a set of input features for developing CBM schemes. The CBM methods can be categorized as $i$) Shallow learning based methods $ii$) Deep learning-based methods $iii$) Hybrid learning-based methods $iv$) Fusion-based methods.

$i)$~\textit{Shallow learning based methods:}
In shallow learning-based CBM schemes, first, a set of features need to be extracted and/or selected and then fed to shallow ML models such as multi-layer NN, RF, SVM, logistic regression, etc. \cite{intro03}.
In \cite{intro02}, authors have extracted Time-Domain (TD) features such as mean, variance, standard deviation, skewness, kurtosis, zero-crossing, etc., followed by feature selection and classification using SVM for bearing fault diagnosis. Similarly, in \cite{intro03}, apart from TD features, authors have also extracted Frequency-Domain (FD) and Time-Frequency (TF) domain features and passed SVM for the classification of faults in air-compressor. Feng \textit{et al.} have presented a review on TF-based features such as empirical mode decomposition, wavelet transform, Wigner-Ville distribution, etc., for CBM of machinery tools \mbox{\cite{intro16}}.
Wu \textit{et al.} \cite{intro19} have presented a fault diagnosis method for steam turbines using principal component analysis and multi-layer NN.
An intelligent diagnosis method based on statistical filter, histogram, and genetic algorithm has been presented for condition monitoring of plant machinery \cite{introrev1}.
In these methods, input features define the performance of classification algorithms. However, knowing what kind of features should be extracted is challenging and empirical. 

$ii)$~\textit{Deep learning based methods:}
Deep learning-based methods learn hierarchical features from raw sensor data through greedy layer-wise training \cite{rev01}. Zhang \textit{et al.} \cite{segmentation} have presented a segmentation method to design input features for Stacked Autoencoder (SAE) based Deep Neural Network (DNN) for bearing fault diagnosis. 
In \cite{rev02}, a single layer Autoencoder (AE) has been used to recognize the faults in induction motors. The input data dimensionality for AE is fixed to $2000$ empirically. Similarly, in \cite{rev03}, authors have presented a fault diagnosis scheme using SAE. Ince \textit{et al.} \cite{rev04} have used 1-D CNN for condition monitoring of the induction motor. In these works, the inputs to AE models are raw segmented data, which are empirically designed inputs. Herein, the input dimensionality is usually high, which may lead to potential concerns such as high computation complexity and overfitting due to massive model parameters. 

$iii)$~\textit{Hybrid learning based methods:}
In hybrid learning-based methods, deep learning models are built upon extracted features from other domains, such as TD, FD, TF, etc. \cite{intro09}.
Thirukovalluru \textit{et al.} \cite{intro10} have extracted TD, FD, and TF features and passed to SAE-based DNN to extract high-level representation followed by the classification of the machine's health states.
A fault diagnosis system has been proposed based on variable beta deep variational autoencoder \cite{intro12}. The 1D data were converted to 2D image data to obtain the inputs for the variational autoencoder. In \cite{intro18}, the sensor signals are partitioned into raw, slope, and curvature data to obtain input features for fault diagnosis.  
In \cite{intro20}, first, sparse dictionary-based input features have been designed and then fed Deep Belief Network (DBN) for bearing fault diagnosis. These methods require domain knowledge and multiple sequential stages to obtain the input features for DNN models.

$iv)$~\textit{Fusion based methods:}
In fusion-based CBM sachems, data level, feature level, and classification level fusion have been performed. In \cite{intro23}, multi-sensor data have been fused for fault classification in ball screw based on extracted manual features and DNN. In \cite{intronew2}, the authors have presented planetary gearbox fault diagnosis using multiple sequential steps for feature extraction followed by fusion of CNN and AE features. Maurya \textit{et al.} \cite{intro11} have fused EMD and DNN-based features for fault diagnosis in rotating machinery components.  
\begin{figure}
\centering
\includegraphics[scale=0.75]{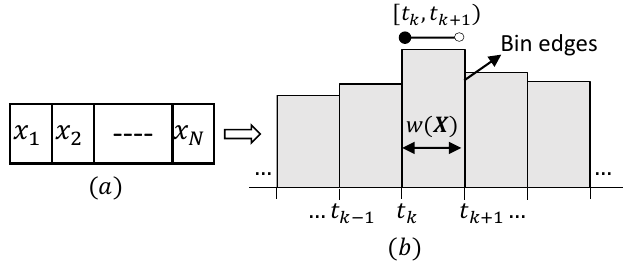}
\caption{Illustration of constructing histogram for a dataset. $(a)$ Dataset: $\textbf{X}= \{x_1, x_2, \cdots, x_N\}$ . $(b)$ Histogram, where $w(\textbf{X})=t_{k+1}-t_k$ denotes bin width having bin interval $[t_k, t_{k+1})$}.
\label{fig: histogram_plot}
\vspace{-3mm}
\end{figure}

In the above-discussed CBM schemes, the common steps are data segmentation followed by feature extraction, feature selection, and/or fusion to obtain the final input feature set for classification using either shallow or deep learning methods. Herein, there are challenges/issues such as data-dependent techniques for data segmentation and input features design, the need for multiple sequential stages and domain knowledge for feature extraction and selection, transformation of data into a different shape (from 1D signal to 2D image), etc. Therefore, this article proposes a novel algorithm to design input features through the lens of histogram theory. Our proposed method unifies the input feature extraction process for different time-series sensor data. The main contributions of this article are summarized as follows.
\begin{enumerate}
    \item A condition-based monitoring scheme based on extracted input features has been proposed. The proposed CBM scheme has been trained end-to-end to recognize the machine's health states using both acoustic as well as vibration datasets. The proposed scheme unifies the feature design process, i.e., applicable to the different applications using different signal signatures. 
    \item This article proposes a novel principled approach for designing input features through the knowledge of histogram theory. A simple feed-forward NN classifier, random forest classifier, and support vector machine classifier are used to validate the effectiveness of extracted input features.
   \item The effectiveness of the proposed CBM scheme has been validated through an acoustic dataset and two vibration datasets, namely, CWRU dataset and IMS-bearing dataset.  

\end{enumerate}  

The remainder of this research article is arranged as follows. Section 1 (Preliminaries on Histogram Theory) highlights the preliminaries for the proposed approach. Section 2 (Proposed Scheme) provides the theory of the proposed scheme for CBM of rotating machinery tools. Two real-time case studies are performed to verify the proposed algorithm in Section 3 (Case Studies, Results and Discussions). Section 4 (Conclusions) concludes this paper.

\section{Preliminaries on Histogram Theory}
\label{Background}
Histograms are a widely practiced statistical tool for summarising and displaying data. In addition, it is a nonparametric density estimator \cite{BGhisto1}. In \cite{BGhisto2}, Scott has proposed optimal and data-based histograms by minimizing integrated mean square error. 
As shown in Figure \mbox{\ref{fig: histogram_plot}}, the histogram can be constructed for a dataset,
$ \textbf{X}= \{x_1, x_2, \cdots, x_N\}\in \mathbb{R}^{N}$, containing $N$ number of data points. The $k$th bin width, $w_k(\textbf{X})=t_{k+1}-t_{k}$, of a histogram with bin interval, $M_k=[t_k, t_{k+1})$,
can be calculated using equation (\mbox{\ref{eqn:hist01}}), where $\sigma$ represents standard deviation \mbox{\cite{BGhisto2}}. Herein, $w_k(\textbf{X})=w(\textbf{X})$, i.e., the bin widths are equal for all $k\in\mathbb{Z}$.
\begin{equation}
\label{eqn:hist01}
w(\textbf{X})=\frac{3.49 \sigma}{N^{1/3}}
\end{equation}

The obtained knowledge from the histogram, i.e., bin width defined in the above equation \mbox{(\ref{eqn:hist01})}, is utilized in the proposed scheme for designing the input features.
\begin{figure}[t]
\centering
\includegraphics[scale=0.72]{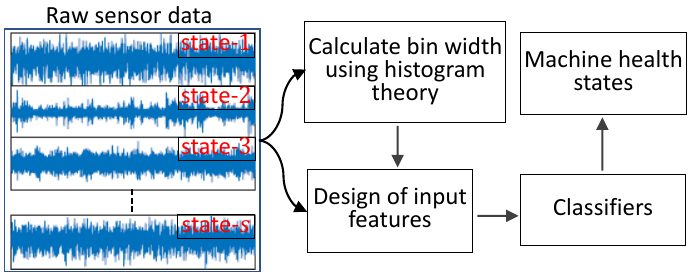}
\caption{The proposed end-to-end framework for condition based monitoring of rotating machines. In this framework, we leverage the histogram theory to determine the appropriate bin width for each state, which is then used to design input features for the corresponding state. These input features are then fed into various classifiers to predict the health state of machines.}
\label{fig: Proposed condition based monitoring framework}
\end{figure}
\begin{figure}[b]
\centering
\includegraphics[scale=0.67]{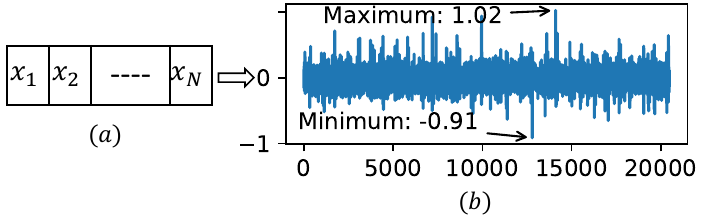}
\caption{Illustration of selecting maximum and minimum amplitudes from the dataset. $(a)$ Data for the $i^{th}$ health state of the machine, i.e., $\textbf{X}^{(i)}= \{x_1, x_2, \cdots, x_N\}$. $(b)$ Plot to depict maximum $(\textbf{X}^{(i)}[max])$ and minimum $(\textbf{X}^{(i))}[min]$ amplitudes.}
\label{fig: minmaxill}
\end{figure}

\section{Proposed Scheme}
\label{sec: Methodology}
This article presents an end-to-end framework for the CBM of machines using the proposed feature extraction scheme. The abstract idea of the proposed scheme is depicted in Figure \ref{fig: Proposed condition based monitoring framework}. 
The raw sensor data are collected through sensors mounted on machinery components. These collected massive sensor data (one-dimensional) are used for extracting input features. For the extraction of input features, knowledge (bin width) obtained using histograms has been utilized. 
The extracted input features are given as input to a feed-forward NN classifier, random forest classifier, and support vector machine classifier to demonstrate the effectiveness of the extracted features.
\begin{algorithm}[t]
\caption{Design of Input Features}
\label{alg:Design of Input Features}
\hspace*{0.56cm}%
\begin{minipage}{0.45\textwidth}%
\textbf{Input:} {$\textbf{S}=\Big\{\textbf{X}^{(1)}, \textbf{X}^{(2)}, \cdots, \textbf{X}^{(s)}   \Big\}$~~// \textit{sensor data for different health states, where $\textbf{X}^{(i)}\in \mathbb{R}^{N_i}$, $ \forall$ $ i=1, 2, \cdots, s$, represent the dataset for $i^{th}$ health state having $N_i$ number of data points.}

\textbf{Output:} $\textbf{F}^{(i)}=\left[\textbf{x}_{1}, \textbf{x}_{2}, \cdots, \textbf{x}_{n_i} \right]^{T}\in\mathbb{R}^{n_i\times m_i}$}, $ \forall$ $ i=1, 2, \cdots, s$~~// \textit{set of output features for $i^{th}$ health state.}
\end{minipage}%
\\

\begin{algorithmic}[1]


\For{$i^{th}$ health state data, i.e., $\textbf{X}^{(i)}$ }
\State\textbf{Initialize:}

\hspace*{0.6cm}%
\begin{minipage}{0.37\textwidth}%
{$x^{(i)}_{min}=\textbf{X}^{(i)}[min]$~and~$x^{(i)}_{max}=\textbf{X}^{(i)}[max]$~~//\textit{initialize with minimum and maximum amplitudes for the $i^{th}$ health state data;}}
\end{minipage}%

 \State Compute $w(\textbf{X}^{(i)})$ using equation (\ref{eqn:hist01})~~// \textit{find the bin width of the histogram for $i^{th}$ health state;} 
 
 \State Compute $m_i$ using equation (\ref{eqn:nooffeatures})~~// \textit{compute the total number of features for $i^{th}$ health state;} 
 
 \State Compute $p^{(i)}=x^{(i)}_{min} + w(\textbf{X}^{(i)})$~// \textit{compute the upper threshold by adding the bin width into the lower threshold $(x^{(i)}_{min}$);}
\For{ all the features $m_i$}
\If{amplitude of $\textbf{X}^{(i)}\geq x^{(i)}_{min}$ and
 \State $\textbf{X}^{(i)} < p^{(i)}$ } 
    \State return $\textbf{x}^{(i)} \subset \textbf{X}^{(i)}$ in $\textbf{F}^{(i)}$~~// \textit{retain the data samples} $\textbf{x}^{(i)}$ \textit{from $\textbf{X}^{(i)}$ for the interval $\big[x_{min}^{(i)}, ~p^{(i)}\big)$, i.e., retain the feature vector as} $\textbf{x}^{(i)}$ from $\textbf{X}^{(i)}$ \textit{if the amplitude is greater than or equal to lower threshold $(x_{min}^{(i)})$ and less than the upper threshold $(p^{(i)})$;}
    \State \textbf{continue;}
    \State $x^{(i)}_{min}= p^{(i)}$~~// \textit{update the lower threshold with the upper threshold;}
    \State $p^{(i)}=x^{(i)}_{min}+w(\textbf{X}^{(i)})$~~// \textit{calculate the new/updated upper threshold;}
\Else
    \If{$p^{(i)}=x^{(i)}_{max}$}
        \State \textbf{break;}~~// \textit{exit from the loop if the upper threshold reaches the maximum amplitude of $\textbf{X}^{(i)}$;}
    \EndIf
\EndIf 
\State \textbf{end}
\EndFor
\State \textbf{end}
\EndFor
\State \textbf{end}
\State \textbf{return} $\textbf{F}^{(i)}$
\end{algorithmic}
\end{algorithm}

\begin{table*}[t]
\centering
\captionsetup{justification=centering}
\caption{States Recognition Performances Using the Proposed Scheme}
\label{tab:proposed result}
\begin{tabular}{c c c c }
\hline \hline
\multirow{2}{*}{\textbf{\makecell{Datasets}}} & \textbf{NN classifier} & \textbf{RF classifier}& \textbf{SVM classifier} \\

 & TPR$/$FPR$/$ACC $\pm$ SD &TPR$/$FPR$/$ACC $\pm$ SD & TPR$/$FPR$/$ACC $\pm$ SD \\ \hline

Acoustic & 0.99$/$0.0001$/$99.96 $\pm$ 0.001 &1.0$/$0.00$/$100.0$\pm$ 0.00  &1.0$/$0.00$/$100.0$\pm$ 0.00  \\ 

CWRU-Vibration & 1.0$/$0.00$/$100.0$\pm$ 0.00 & 1.0$/$0.00$/$100.0$\pm$ 0.00  & 1.0$/$0.00$/$100.0$\pm$ 0.00  \\ 
IMS-Vibration & 1.0$/$0.00$/$100.0$\pm$ 0.00 & 1.0$/$0.00$/$100.0$\pm$ 0.00  & 1.0$/$0.00$/$100.0$\pm$ 0.00  \\ 
  \hline\hline
\end{tabular}
\end{table*}

\begin{figure*}
\centering
\begin{minipage}{.5\textwidth}
  \centering
  \includegraphics[width=1\linewidth]{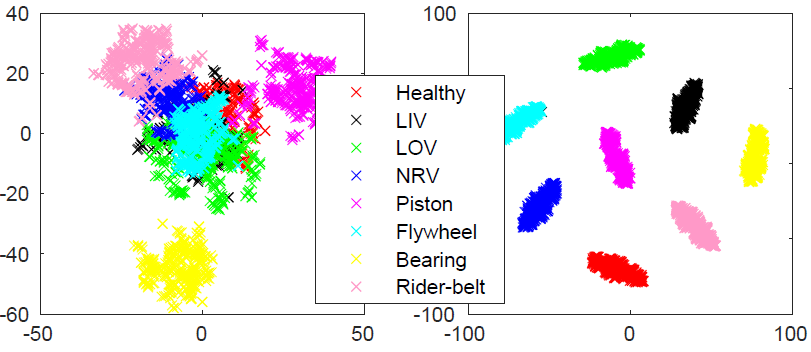}
  \captionof{figure}{Visualization of features for \textbf{acoustic} dataset. \textbf{Left:} raw sensor data and \textbf{Right:} Extracted features using the proposed algorithm.}
  \label{fig:scatterair}
\end{minipage}~~~~~
\begin{minipage}{.5\textwidth}
  \centering
  \includegraphics[width=0.95\linewidth]{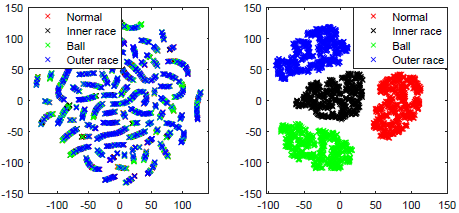}
  \captionof{figure}{Visualization of features for \textbf{vibration} dataset. \textbf{Left:} raw sensor data and \textbf{Right:} Extracted features using the proposed algorithm.}
  \label{fig:scattervib}
\end{minipage}
\end{figure*}

\subsection{Design of Input Features}
\label{subsec:Design of Input Features}
A novel method for designing input features has been proposed to get the training samples from sensor data that contain massive data points. The proposed scheme unifies the feature design process, i.e., applicable to the different applications using different signal signatures. 

The design of input features for the time-series sensor data collected from the rotary machines in different health conditions (healthy and various faulty conditions) has been performed as follows. Let, $\textbf{X}^{(i)}\in \mathbb{R}^{N_i}$ $\forall$ $ i=1, 2, \cdots, s$, represent the data for the $i^{th}$ health state of the machine, where $s$ is the total number of health states and $N_i$ is the total number of data points in the $i^{th}$ health state. The number of designed features or the dimensionality of the designed feature for the $i^{th}$ health state, denoted by $m_i$, can be calculated using the following equation \mbox{(\ref{eqn:nooffeatures})}.
\begin{equation}
\label{eqn:nooffeatures}
m_i=\frac{\textbf{X}^{(i)}[max]-\textbf{X}^{(i)}[min]}{w(\textbf{X}^{(i)})}
\end{equation}
Where $\textbf{X}^{(i)}[max]$ represents the data point having the maximum amplitude in the $i^{th}$ health state data and, similarly, $\textbf{X}^{(i)}[min]$ represents the data point with the minimum amplitude. These points are illustrated in Figure \mbox{\ref{fig: minmaxill}}; as shown in this Figure, the maximum and minimum amplitudes are $1.02$ and $-0.91$, respectively. The term $w(\textbf{X}^{(i)})$ is the bin width for the $i^{th}$ health state data that is calculated using equation \mbox{(\ref{eqn:hist01})}, discussed above in Section 1 (Preliminaries on Histogram Theory). 
The feature extraction process is given in Algorithm \mbox{\ref{alg:Design of Input Features}} and discussed follows. The inputs to the algorithm are the raw sensor data for different health states, i.e., $\textbf{X}^{(1)}=\big\{x_1, x_2, \cdots, x_{N_1} \big\}$, $\textbf{X}^{(2)}=\big\{x_1, x_2, \cdots, x_{N_2} \big\}$, $\cdots$, and $\textbf{X}^{(s)}=\big\{x_1, x_2, \cdots, x_{N_s} \big\}$. For $i^{th}$ health state data, the first step is to initialize the $x_{min}^{(i)}$ and $x_{max}^{(i)}$ with the minimum and maximum amplitude of $\textbf{X}^{(i)}$, respectively. Then using the bin width (given equation \mbox{(\ref{eqn:hist01})}), the dimensionality of the feature vector, $m_i$, for $i^{th}$ health state is calculated using equation \mbox{(\ref{eqn:nooffeatures})}. After this, in step 5 of Algorithm \mbox{\ref{alg:Design of Input Features}}, $p^{(i)}=x^{(i)}_{min} + w(\textbf{X}^{(i)})$ is calculated in order to get the number of samples within the amplitude interval $\big[x_{min}^{(i)}, ~p^{(i)}\big)$. Consequently, from steps 6 to 9 of the algorithm,  the feature vector $\textbf{x}^{(i)}$ is obtained. The length of $\textbf{x}^{(i)}$, i.e., the number of data points/samples, denoted by $n_i$, can be obtained using equation (\mbox{\ref{eqn:hist1}}).
\begin{equation}
\label{eqn:hist1}
n_i=\sum_{j=1}^{N_i}I^{(i)}(x_j)~~\forall~~ i=1, 2, \cdots, s
\end{equation}

\begin{equation}
\label{eqn:hist2}
 I^{(i)}(x_j) =
    \begin{cases}
      1 & \text{if}~~x_j \in \big[x_{min}^{(i)}, ~p^{(i)}\big) \\
      0 & \text{otherwise} 
    \end{cases}
\end{equation}
Where $I$ is an indicator function, defined in the above equation (\mbox{\ref{eqn:hist2})}. The obtained length of $\textbf{x}^{(i)}$ will always be less than the length of $\textbf{X}^{(i)}$ because $\textbf{x}^{(i)}$ is the proper subset of $\textbf{X}^{(i)}$, i.e., $\textbf{x}^{(i)}$ is obtained from  $\textbf{X}^{(i)}$ only for the amplitude interval $\big[x_{min}^{(i)}, ~p^{(i)}\big)$.
Further, from steps 10 to 15 of the algorithm, the process is continued for all number of features until the upper threshold reaches the maximum amplitude of $\textbf{X}^{(i)}$. Herein (steps 11 and 12 of the algorithm), the lower threshold is updated with the upper threshold, and the upper threshold is updated by adding the bin width to the newly updated lower threshold.
Finally, the designed input feature space, $\textbf{F}^{(i)} \in \mathbb{R}^{n_i \times m_i}$, for the $i^{th}$ health state data is given in the following equation (\ref{eqn:inputfeaturesall}), where $\textbf{x}_j^{(i)}\in \mathbb{R}^{m_i}$. 
\begin{equation}
\label{eqn:inputfeaturesall}
\textbf{F}^{(i)}=\big[ \textbf{x}^{(i)}_1,\textbf{x}^{(i)}_2 \cdots \textbf{x}^{(i)}_{n_i} \big]^{T}
\end{equation}
\subsection{State Recognition}
\label{State Recognition1}
The obtained feature space ($\textbf{F}^{(i)}$) is fed to the classifiers for predicting the states of the machines. Feed-forward NN classifier \cite{nn1}, Random Forest (RF) classifier \cite{s4}, and Support Vector Machine (SVM) classifier \cite{s2} have been used for the generalization of the proposed algorithm.
\begin{table*}[t]
\centering
\captionsetup{justification=centering}
\caption{Comparison of Accuracy (in $\%$) of Proposed Scheme with state-of-the-art Methodologies}
\label{tab:Comparison}
\resizebox{0.677\textheight}{!}{%
\begin{tabular}{l| c c c| c c c|c c c}
\hline \hline
\multirow{2}{*}{\textbf{~~~~~~~~Methods}}& \multicolumn{3}{c|}{\textbf{\textbf{Acoustic Dataset}}} & \multicolumn{3}{c|}{\textbf{\textbf{CWRU Vibration Dataset}}} &\multicolumn{3}{c}{\textbf{\textbf{IMS Vibration Dataset}}}\\ \cline{2-10}
&NN-CLF  & RF-CLF&SVM-CLF &NN-CLF  & RF-CLF&SVM-CLF& NN-CLF  & RF-CLF&SVM-CLF \\ \hline
\textbf{Proposed} &\textbf{99.96}&\textbf{100.0}&\textbf{100.0}&  \textbf{100.0}&\textbf{100.0}&\textbf{100.0}& \textbf{100.0}&\textbf{100.0}&\textbf{100.0}\\
TD Features \cite{intro02}&93.47& 92.67&93.00 &98.88&99.00&99.01& 65.97&68.77&71.87\\
FD Features \cite{intro10}&99.56&99.67& 99.79  &89.78&89.67&89.67&95.69&96.77&96.97\\
DNN Features \cite{segmentation}&98.72&98.56 &98.87&87.50&87.56&88.00&93.72&94.06&95.12\\
Features Fusion \cite{intro11}&97.18&95.78 &96.59 &99.57&99.93&99.48&67.08&69.72&70.50\\
Multi Features \cite{intro03}&97.56&98.92 &98.97  &93.39&9.01&94.00&96.50&97.02&97.91\\
Segmented Raw Input \cite{segmentation}&68.33& 70.20&73.32& 64.40&67.56&68.00& 63.33&61.20&67.32\\
\hline
\hline 
\end{tabular}
}
\hspace{150ex}*\textit{NN-CLF, RF-CLF, and SVM-CLF represent Neural Network Classifier, Random Forest Classifier, and Support Vector Machine Classifier, respectively.}
\end{table*}
\section{Case Studies, Results, and Discussions}
\label{sec: Case Studies, Results}
Three case studies for the CBM of machines are included to validate the performance of the proposed scheme. All the computations are performed on a computer with $16$ GB RAM and $\mathtt{i}7$-$8700$ processor.
\begin{figure}[t]
\centering
\includegraphics[scale=0.82]{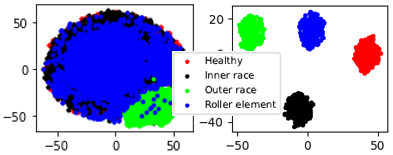}
\caption{Visualization of features for IMS vibration dataset. \textbf{Left:} raw sensor data and \textbf{Right:} Extracted features using the proposed algorithm.}
\label{fig: Ims_tsne}
\end{figure}
\subsection{Evaluation Metrics}
\label{Evaluation Metrics}
The performance of the proposed scheme is measured by True Positive Rate (TPR), False Positive Rate (FPR), and average accuracy (ACC) of the $\mathtt{k}$-folds Cross-Validation (CV). Standard Deviation (SD) in the accuracies within folds has also been calculated 
using (\ref{SD}), where $\mathtt{sd(\cdot)}$ and $\mathtt{k}$ denote the standard deviation operator and the number of folds, respectively. 
\begin{equation}
\label{SD}
  \text{SD}=  \frac{1}{\sqrt{\mathtt{k}}}\mathtt{sd}(\text{ACC}_1, \text{ACC}_2, \cdots, \text{ACC}_{\mathtt{k}})
\end{equation}

\subsection{Data Description}
\label{sec: Data Description}
\subsubsection{Acoustic Dataset:}
\label{subsec: Acoustic Dataset}
The real-time acoustic data \cite{intro03} has been acquired from a reciprocating type air compressor. The dataset contains eight states, including a \textit{healthy} state and seven \textit{faulty} states: \textit{Leakage Inlet Valve (LIV),  Leakage Outlet Valve (LOV), Non-Return Valve (NRV), piston, flywheel, bearing}, and \textit{rider-belt}.
For each state of the air compressor, $225$ recordings are collected for $5$ seconds at the sampling rate of $50$ kHz. 
\subsubsection{CWRU Vibration Dataset:}
\label{subsec: Vibration Dataset}
The vibration datasets \cite{intro25} were collected by Case Western Reserve University (CWRU) bearing data center at the sampling rate of $12$ kHz. The test bearing has single point seeded faults of diameters of $0.007$, $0.014$, $0.021$, $0.028$, and $0.040$ inches, which supports the motor shaft. 
The used datasets contain a \textit{normal} condition and three \textit{faulty} conditions (inner race, ball, and outer race) with $0.014$ inches of faults diameter at the motor speed of $1750$ RPM.
\subsubsection{IMS Vibration Dataset:}
\label{subsec: IMS Dataset}
The Intelligent Maintenance System (IMS) bearing dataset \mbox{\cite{Imsdata}} is provided by the University of Cincinnati. Three sets of data were collected from a run-to-failure experiment involving four bearings on a loaded shaft. The load on the shaft was $6000$ pound-mass, and it was rotating at the speed of $2000$ Revolutions Per Minute (RPM). To collect data from each bearing, high-precision accelerometers were installed. For each bearing, the dataset comprises vibration signals for 1 second, with $2156$, $984$, and $4448$ files for set-1, set-2, and set-3 datasets, respectively. The vibration signal is sampled at the rate of $20$KHz. The dataset contains four health conditions: healthy and three types of defects, namely, inner race, outer race, and roller element.
\subsection{Results and Discussions}
\label{sec: Results and discussions1}
The acoustic dataset has $2.5\times 10^5$ data points (one-dimensional) in each recording, as discussed in \textit{Data Description Section}. The proposed feature extraction method is applied to get the training samples. The obtained number of features and training samples for acoustic dataset are $54$ and $1400$, respectively, using Algorithm \ref{alg:Design of Input Features}. The CWRU vibration dataset has at least $1.2\times 10^5$ data points in every bearing condition; the obtained features and training samples are $34$ and $1000$, respectively, using the proposed feature extraction method. Similarly, the IMS dataset has an order of $10^8$ data points (one-dimensional) for each health condition. Using the proposed Algorithm, the obtained number of features and training samples for the IMS dataset are $11$ and $2200$, respectively.

A visualization algorithm called \enquote{t-SNE} \cite{tsne} is implemented to demonstrate the effectiveness of the features extracted by the proposed algorithm. The \enquote{t-SNE} algorithm converts high-dimensional features into two-dimensional maps to visualize the separation of the classes. The resulting two-dimensional maps for the acoustic dataset are plotted as a scatter-plot in Fig. \ref{fig:scatterair}. Similarly, for the CWRU vibration dataset and IMS vibration dataset, the maps are in plotted Fig. \ref{fig:scattervib} and Fig. \mbox{\ref{fig: Ims_tsne}}, respectively.
In Fig. \ref{fig:scatterair}, Fig. \ref{fig:scattervib}, and Fig. \ref{fig: Ims_tsne}, the left figures show the scatter-plot of raw sensor data, and the right figures show the scatter-plot of extracted features using the proposed scheme. It can be observed that the extracted features using the proposed scheme for different states are well separated compared to raw sensor data for health state recognition. 

Multi-layer NN, RF, and SVM classifiers have been used for fault classification for all acoustic, CWRU vibration, and IMS vibration datasets. The NN classifier is designed with $60$ and $20$ neurons in the first and second hidden layers, respectively, and $8$ neurons in the output layer for acoustic data. For CWRU vibration, $40$ and $15$ neurons are used in the first and second hidden layers, respectively, and $4$ neurons are in the output layer. Similarly, $30$ and $15$ neurons are fixed in the first and second hidden layers, and $4$ neurons are in the output layer for IMS vibration data. Adam optimizer is used to train the networks.
The SVM classifier is implemented using LIBSVM package\cite{libsvmm}. The RF model is implemented using the TreeBagger class of Statistics and Machine Learning library.
The average classification accuracy (in \%) and standard deviation with $5$-fold CV are given in Table \ref{tab:proposed result}. 
TPR and FPR are also calculated to show the effectiveness of the designed features. The closer the value of TPR and ACC to $1$ and FPR to $0$, indicate better performance.




The performance of the proposed scheme has been compared with the state-of-the-art methods. In \cite{intro02}, TD features are extracted from segmented sensor data. In \cite{intro10}, FD input features are designed for faults classification.  In \cite{segmentation}, raw sensor data has been segmented to design input, followed by DNN to extract deep features for fault classification.  Low-level and high-level representations are fused for fault diagnosis using binary classifiers \cite{intro11}.
Multi-statistical features are stacked to form an input set for classifiers to recognize the health states \cite{intro03}.
In comparison, the proposed scheme has performed the best across all the datasets, as shown in Table \ref{tab:Comparison}. Using the RF and SVM classifiers, the highest average accuracy with the proposed method is $100\%$ for each dataset. Similarly, the height accuracy of the NN classifier is $99.96\%$ for acoustic data and $100\%$ for both vibration datasets. In general, it can be observed that the proposed scheme has shown satisfactory improvement in most cases. 
\section{Conclusions}
\label{sec: Conclusions}
This paper proposes an end-to-end solution for the CBM of rotating machines to unify the feature extraction methods for different signal signatures. This research article build-up a new method of extracting input features over the empirical and conventional approaches. The proposed scheme removes the need for extracting low-level features and multiple pre-processing steps in CBM systems for real-time applications. The demonstrations of the effectiveness of the proposed method through three case studies show the broad applicability for the CBM of machinery components. 

The future work will include solving the imbalanced problem through deep generative models and the fusion of different sensor data to enhance the applicability of the presented methods.

\ifCLASSOPTIONcaptionsoff
  \newpage
\fi

\bibliographystyle{plain}
\bibliography{Reference.bib}

\end{document}